\documentclass[conference]{IEEEtran}
\IEEEoverridecommandlockouts
\usepackage{cite}
\usepackage{amsmath,amssymb,amsfonts}
\usepackage{algorithmic}
\usepackage{graphicx}
\usepackage{textcomp}
\usepackage{xcolor}
\usepackage{comment}
\usepackage{scalerel,stackengine}
\def\BibTeX{{\rm B\kern-.05em{\sc i\kern-.025em b}\kern-.08em
    T\kern-.1667em\lower.7ex\hbox{E}\kern-.125emX}}
\ifCLASSINFOpdf

\else
 
\fi

\usepackage{eso-pic}
\newcommand\AtPageUpperMyright[1]{\AtPageUpperLeft{%
 \put(\LenToUnit{0.5\paperwidth},\LenToUnit{-1cm}){%
     \parbox{0.5\textwidth}{\raggedleft\fontsize{9}{11}\selectfont #1}}%
 }}%
\newcommand{\conf}[1]{%
\AddToShipoutPictureBG*{%
\AtPageUpperMyright{#1}
}
}
\begin{document}

\conf{2019 22nd International Conference on Computer and Information Technology (ICCIT), 18-20 December, 2019}

\title{Authorship Attribution in Bangla literature using Character-level CNN\\}

\author{
\IEEEauthorblockN{Aisha Khatun}
\IEEEauthorblockA{Department of Computer \\Science and Engineering\\
Shahjalal University of\\
Science and Technology\\Sylhet, Bangladesh\\
Email: aysha.kamal7@gmail.com}
\and
\IEEEauthorblockN{Anisur Rahman}
\IEEEauthorblockA{Department of Computer \\Science and Engineering\\
Shahjalal University of\\
Science and Technology\\Sylhet, Bangladesh\\
Email: emailforanis@gmail.com}
\and
\IEEEauthorblockN{Md. Saiful Islam}
\IEEEauthorblockA{Department of Computer \\Science and Engineering\\
Shahjalal University of\\
Science and Technology\\Sylhet, Bangladesh\\
Email: saif.acm@gmail.com}
\and
\IEEEauthorblockN{Marium-E-Jannat}
\IEEEauthorblockA{Department of Computer \\Science and Engineering\\
Shahjalal University of\\
Science and Technology\\Sylhet, Bangladesh\\
Email: jannat-cse@sust.edu}}

\IEEEoverridecommandlockouts
\IEEEpubid{\begin{minipage}[t]{\textwidth}\ \\[10pt]
        \centering\normalsize{978-1-7281-5842-6/19/\$31.00~\copyright2019 IEEE}
\end{minipage}}

\maketitle

\begin{abstract}
Characters are the smallest unit of text that can extract stylometric signals to determine the author of a text. In this paper, we investigate the effectiveness of character-level signals in Authorship Attribution of Bangla Literature and show that the results are promising but improvable. The time and memory efficiency of the proposed model is much higher than the word level counterparts but accuracy is 2-5\% less than the best performing word-level models. Comparison of various word-based models is performed and shown that the proposed model performs increasingly better with larger datasets. We also analyze the effect of pre-training character embedding of diverse Bangla character set in authorship attribution. It is seen that the performance is improved by up to 10\% on pre-training. We used 2 datasets from 6 to 14 authors, balancing them before training and compare the results.
\end{abstract}
\hfill

\section{Introduction}

Authorship attribution is generally concerned with the identification of the original author of a given text from a set of given authors. It has a wide range of applications including plagiarism detection, forensic linguistics, etc. Each author has a distinctive writing style that is exploited by statistical analysis to detect the author.

However, in Bangla language, the amount of work done in this area is not very rich despite being one of the most spoken languages. In traditional methods, texts are represented using independent features such as lexical n-gram or frequency-based representation. In this approach, words of similar context are likely to be represented in different vector space as the features are independent. So, the semantic values of the words might be lost, which is problematic. Word embedding, also generally known as distributed term representations, offers a solution to this problem by encoding semantic similarity from their co-occurrences. Chowdhury\cite{bhai} experimented with the effectiveness of word embedding in authorship attribution for Bangla language for various architectures.\newline

Another type of embedding, which we tried to analyze in this paper is character embedding. Character CNN was first introduced by Zhang\cite{zhang} for the text classification task. Through the empirical experiment of Sebastian \cite{ruder2016character} and Jozefowicz\cite{googlebrain}, character level NLP has been proven to be very promising in various ways. Although it may seem that character on its own does not have any semantic value, Radford \cite{radford2017learning} illustrates that character-level models can capture the semantic properties of text. Character level models are also better at handling out-of-vocabulary words, misspelling, etc and provide an open vocabulary. Another major advantage is that it reduces the dimension to as low as 16, unlike word embedding where the dimension can increase up to 300 while the vocabulary is also huge. So, character embedding removes the bottleneck in training tasks and gives huge advantages on computational complexity.\newline

Our approach in this paper was to investigate how
character embedding performs in the task of Authorship Attribution in Bangla language. Bangla Language has numerous words with joint letters which can be written in a few different forms. Moreover, there are some words with the same meaning but slightly different spelling. These inconsistencies are not recognized by word-level models but character-level models can capture and relate words of this kind, making such models more appropriate for Bangla language. Comparison of character embedding with word embedding is discussed according to the findings. Experiments with and without pre-trained embedding layers have also been done to show the effectiveness of information captured in the embeddings.
No previous work, analysis or investigation has yet published on the effect of character embedding in Authorship Attribution of Bangla Literature as of our knowledge to date.
This paper follows the structure provided below:
\begin{itemize}
    \item Related Works - Extensive background study on some works relevant to this paper are provided in this section. 
    \item Corpus - The dataset used in our experiment is described in this section.
    \item Methodology - The proposed architecture for our character embedding model along with the strategies used during the training phase of the neural networks are described in depth.
    \item Experiments - Describes the evaluation process and the model setup for comparison.
    \item Results and Discussion - Our findings along with results and possible reasons are presented in this part.
    \item Conclusion - In the last section, some recommendations and scope for future research on this field are mentioned.
\end{itemize}
\section{Related Works} \label{related}

\subsection{On Authorship Attribution}
Authorship Attribution has been a topic of important research for a long time. With increased anonymity on the internet and easy fraud, authorship attribution of writings has become crucial. For authorship attribution, work on varying degrees of feature selection\cite{stamatatos2009survey},
including advanced features such as local histograms\cite{escalante2011local}. Naive similarity-based models\cite{koppel2011authorship}, SVMs\cite{narayanan2012feasibility} have been explored. Semi-supervised approach to authorship attribution was also taken\cite{nasir2014off}. SOTA was achieved by Ruder\cite{ruder2016character} using character-level and multi-channel CNN.\newline

Compared to other works, very few works have been done in Bangla language, lacking any sort of high benchmarks until very recently. Das and Mitra\cite{das2011author} worked with a really small dataset of 36 documents and 3 authors to perform uni-gram and bi-gram feature-based classification. Chakraborty\cite{chakraborty2012authorship} worked with SVMs on 3 authors to achieve up to 84\% accuracy. Shanta Phani also attempted to attribute 3 authors with machine learning\cite{phani2015authorship}. P. Das, R. Tasmim, and S. Ismail used 4 authors of current times and hand-drawn features such as word frequency, type-token ratio, number of various POS, word/sentence lengths etc\cite{das2015experimental}. 90.67\% was achieved by Hossain and Rahman by using multiple features along with cosine similarity\cite{hossain2017stylometric}. Pal, Siddika, and Ismail achieved 90.74\% accuracy with 6 authors using SVM on one feature\cite{pal2017machine}. Multi-layered perceptrons were employed by Phani, Lahiri, and Biswas\cite{phani2016machine}. Impressive results were achieved very recently by\cite{bhai} using various word embeddings on a 6 author dataset. They demonstrated the effects of various architectures and word embeddings on authorship attribution and concluded that fastTexts skip-gram used with CNN tends to beat all other models in terms of accuracy. No work has been done on the character level classification task as of knowledge in Bangla literature. The effects of Bangla alphabet complexity and language formulation on architectural design and character embedding learning remains largely untouched.

\subsection{On Embedding}
Embeddings are effectively mappings from various entities (character, word, sentence, etc) to continuous vector spaces in high dimensions. The relation among the numerical representations gives a semantic, syntactic and morphological meaning of the entities. These meanings are leveraged by machine learning techniques to find patterns in texts and thus perform various tasks such as classification.\newline

\subsubsection{Word Embedding}
Representing words in continuous vector spaces is considered as one of the breakthroughs of NLP. Word embeddings are learned in the form of an embedding layer or separately in an unsupervised manner. Among the unsupervised techniques include Continuous Bag-of-Words(CBOW) and Skip-Gram models famously implemented by Word2Vec and fastText. Also, there are co-occurrence statistical methods such as Glove.
Santos\cite{santos2017sentiment} used word embeddings with convolutional models showing significant improvements over baseline methods. Word embeddings have been used to improve the performance of sentiment analysis\cite{rudkowsky2018more}. Often pre-trained embeddings are used or are learned for specific tasks such as tree-structured long short-term memory networks\cite{tai2015improved} and Multi-perspective sentence similarity modeling\cite{he2015multi}. Although words started to be used as units of text, various works have started to break down words and work at subword and character levels. Wieting\cite{wieting2016charagram} creates subword embedding from counts of character n-grams.\newline

\subsubsection{Character Embedding}

Character Level embeddings are used in various ways, either by themselves or to produce embeddings of higher levels e.g for words. Character embeddings have been employed in POS tagging\cite{ling2015finding}, language modelling\cite{ling2015finding} and dependency parsing\cite{ballesteros2015improved}. Character-RNN were used for machine translation, for representing words\cite{luong2016achieving} or to generate character level translations\cite{chung2016character}. Pure Character level classification was first explored using CNN architecture\cite{zhang}. Jozefowicz\cite{googlebrain} shows that a character-level language model can significantly outperform state of the art models. Their best performing model combines an LSTM with CNN input over the characters. Besides using either just word or character embeddings, ideas of combining them also have been introduced\cite{liang2017combining}. Attempts to learn character embedding and serve as pre-trained have also been explored\cite{char2vec}. \newline

\section{Corpus}
\label{corpus}

Because of the scarcity for the standard dataset in authorship attribution, we made a custom web crawler to parse the data on our own. We collected writings from an online Bangla e-library containing writings(e.g., novels, story, series, etc.) of different authors. Table \ref{corpustable} shows the details of our dataset. Our dataset is larger compared to the previously worked on datasets for Bangla as mentioned in section \ref{related} with 13.4+ million words. The dataset was equally partitioned with each document having the same length of 750 words. Various subsets of authors were chosen and the dataset was truncated to each author having the same number of samples.

The dataset from the paper\cite{bhai} was also used. This dataset consists of 6 authors with 350 sample texts per author and total word count of 2.3+ million.\newline

\begin{table}[h!]
\centering
\caption{Corpus details}
\begin{tabular}{|c| c c |} 
\hline
Author & Word count & Unique words\\
\hline\hline
candidate 01 & 351750 & 44477\\
\hline
candidate 02 & 421500 & 62485\\
\hline
candidate 03 & 825000 & 53163\\
\hline
candidate 04 & 666000 & 84888\\
\hline
candidate 05 & 636750 & 67579\\
\hline
candidate 06 & 984000 & 78717\\
\hline
candidate 07 & 944250 & 89956\\
\hline
candidate 08 & 3388500 & 161893\\
\hline
candidate 09 & 357000 & 43864\\
\hline
candidate 10 & 786000 & 69182\\
\hline
candidate 11 & 1056000 & 69648\\
\hline
candidate 12 & 1472250 & 109230\\
\hline
candidate 13 & 698250 & 76071\\
\hline
candidate 14 & 581250 & 84311\\
\hline

\end{tabular}
\break
\label{corpustable}
\end{table}

For pre-training our model, we used another large corpus of  Bangla Newspaper articles based on 6  topics. The topics were accident, crime, education, entertainment, environment, and sports. The dataset consists of 10564543 tokens.
\section{Methodology}

\subsection{Proposed Architecture}

Character-level CNN can sufficiently replace words for classifications\cite{zhang}. This means CNN does not require the syntactic or semantic structure of a language, which makes such approaches effectively independent of language as the number of characters is limited. To this end, CNN was used in this paper to perform the task of author attribution. An elaborate set of experiments were performed on 3 different datasets to conclude with an architecture that successfully extracts the character level features of any sample text. The same architecture was used to prepare the pre-trained character embeddings for classification tasks. The model is a deep neural network starting with 4 convolutional layers, each followed with a maxpool layer of kernel size 3. As standardized in computer vision, for the convolutional layers, the number of filters increases while decreasing the kernel size at each layer. The kernel sizes are respectively 7,3,1 and 1. The number of filters is 64,128,256 and 256. Beneath all is an embedding layer where each character is represented as a vector of length $\|V\|$, i.e, the alphabet size. The convolutional layers are stacked with a fully connected layer of 512 activation nodes, activation function ReLU and dropout. Finally, an output layer with softmax is used to provide the classification probabilities. For optimization Adam optimizer is used along with categorical cross-entropy as the loss function.

\subsection{Character Embedding}

Character embedding aims to turn characters into meaningful numerical representations in the form of vectors. These vectors may represent the correlation of different characters, or even correlation of groups of characters together i.e. words, sentences, documents, etc. This concept can be leveraged to use character embeddings to fit misspelled words, rare or new words, slangs or emoticons. They can also easily represent words with variations such as drive, driving, drives, etc. There is no more bottleneck for out of vocabulary words. The character set can be used to make any word, even if it is out of vocabulary, in contrast to word embeddings which simply ignored them, or had weak representations for rare words. This way character embeddings increase generalization compared to words. Another significant improvement is the vocabulary size. Instead of having a very large vocabulary of words, character embeddings have a fixed number of characters which is significantly smaller, therefore reduces model complexity and the number of parameters by a significant amount. Furthermore, they can be represented with a small vector size (e.g 16) and still be significantly informative as opposed to word embeddings which require at least 100-300 size vectors for a decent model. The simplest way to represent a character is to use a one-hot encoding. This requires the vector size to be the size of the alphabet. We used a one-hot encoding as a baseline for comparison of pre-trained embeddings. Otherwise, one can randomly initialize the vectors, where the vectors can be of any size as small as 16 to as big as 300. This becomes a hyperparameter for tuning.

\subsection{Training the model}

The alphabet size, and therefore the embedding vector size is 253. Among the 253 different characters are the English letters(capital and small) and digits, Bangla letters and digits, Bangla vowel symbols, and various other punctuation and symbols. For comparative training, two sets of embeddings were created for the character set. First is one-hot encoding, and the other is pre-trained embeddings. The training was done in two phases as stated below:\newline

\subsubsection{Pre-training Embedding}

To learn character embeddings, the architecture mentioned above was used for classification of the news dataset as mentioned in section \ref{corpus}. This is in contrast to the usual ways of learning embeddings. No separate model was used\cite{char2vec} to learn the embeddings. Instead, already available classification task on a marginally large dataset learns character embeddings for its purposes. These embeddings can be used as initialization for the author attribution task, which has a smaller dataset compared to the former, giving it an initial boost. The model was trained with a learning rate of 0.001 and decay of 0.0001. The maximum length of each text sample was set as 1000 and batch size as 80. A dropout rate of 0.5 was used in the fully connected layer to prevent over-fitting. The embeddings then learned to have an understanding of how the Bangla language works and provide a meaningful initialization for any classification tasks. They were then extracted and used for the task of authorship attribution.\newline

\subsubsection{Performing Classification}

To perform the main task of author attribution and comparison, this training phase was performed twice with each type of embeddings mentioned above, i.e one-hot and pre-trained. The fully connected layer was given a dropout probability of 0.7 and trained with batch size 128 and the maximum length of each text was set to be 3000 characters. Everything else was kept similar. The classification was carried out with 2 author attribution datasets, one with 6 authors\cite{bhai} and our dataset with maximum 14 authors. The larger dataset was trained with 6,8,10,12 and 14 authors to analyze the effects of increasing classes on the proposed model.

\section{Experiments}

We evaluate the performance of the proposed architecture in terms of accuracy, with and without pre-training character level embedding and comparing them on the held-out dataset. We also try to infer how the character-level model compares with the word level models. All models are compared for the increasing number of authors(classes) on the corpus mentioned to assess the quality of the models. To keep the dataset balanced, the number of samples per class were truncated to the minimum among the classes. We propose a model for word-level classification mostly similar to our Char-CNN model. The model used for performance analysis is as follows:\newline




\subsection{Word Embedding Model}

This model has a close resemblance to the proposed Char-CNN model except for a few differences to tune with the word level version of the classification. The model has 2 convolutional layers with the kernel sizes 7,3 and number of filters are  128,256 respectively for each layer. Each layer followed by a maxpool layer. The model is initialized with pre-trained word embeddings from word2vec and fastText, both CBOW and skip-gram versions. The convolutional layers are stacked with an LSTM layer of 100 neurons and a fully connected layer of 512 activation nodes both with dropout to prevent overfitting. Finally, a softmax layer is used to provide the classification probabilities. It is trained for 10 epochs with a learning rate of 0.001 with Adam optimizer, the batch size is 32 and 750 words per sample are used as input to models. All the word level models have a vocabulary size of 60000 and word embedding vector of size 300.

\section{Results and Discussion}

The accuracies achieved(in percents) on the test set of the datasets, with pre-trained embeddings for both word and character levels are summarized in Table \ref{resulttable}. Because the datasets were balanced, the comparison of accuracies is sufficient.

\begin{table}[h!]
\centering
\caption{Performance comparison of different models with pre-trained embedding}
\begin{tabular}{|c| c c c c c c|} 
\hline
\#of Authors & 6\cite{bhai} & 6 & 8 & 10 & 12 & 14\\ 
samples/author & 350 & 1100 & 931 & 849 & 562 & 469\\
\hline\hline
Char-CNN & 83 & 96 & 92 & 86 & 75 & 69\\
\hline
W2V(CBOW) & 65.3 & 97 & 82.8 & 83.3 & 76.4 & 71.8\\
\hline
fastText(CBOW) & 65 & 73 & 58 & 35.7 & 37.31 & 40.3\\
\hline
W2V(Skip) & 79 & 94 & 91.1 & 85.4 & \textbf{82.2} & 78.6\\
\hline
fastText(Skip) & \textbf{86} & \textbf{98} & \textbf{95.2} & \textbf{86.35} & 80.9 & \textbf{81.2}\\
\hline
\end{tabular}
\break
\label{resulttable}
\end{table}

Accuracy comparison(in percents) of the proposed model with and without pre-trained character embeddings are summarized in Table \ref{comptable}.

\begin{table}[h!]
\centering
\caption{pretrained vs non-pretrained comparison}
\begin{tabular}{| c | c c c c c c |} 
 \hline
 \#of Authors & 6\cite{bhai} & 6 & 8 & 10 & 12 & 14\\ [0.5ex] 
 \#of samples/class & 350 & 1100 & 931 & 849 & 562 & 469\\
 \hline\hline
 Pretrained Embedding & 83 & 96 & 92 & 86 & 75 & 69\\
 \hline
 Not pretrained & 71 & 95 & 82 & 83 & 66 & 59.5\\
 \hline
 
\end{tabular}
\break
\label{comptable}
\end{table}

\begin{figure}[h!]
\includegraphics[scale=0.555]{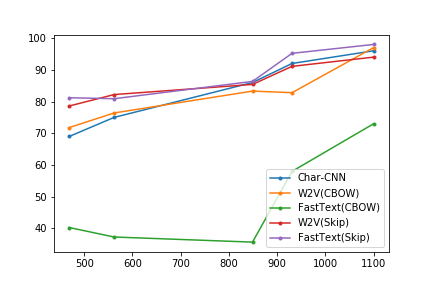}
\centering
\caption{Accuracy of various models with increasing number of samples.}
\label{Fig:comparison}
\end{figure}

From the accuracy comparisons shown in Table \ref{resulttable} we see that Skip-gram implemented by fastText performs well in the given datasets. So we can infer that subword level classification tends to extract a good amount of meaning information and styles from the text. On the other hand, the word2vec models, which use entire words have worse performance. Character level model performs reasonably well in competition with subword level as long as the dataset is big enough. When the number of authors increased, the number of samples per author decreased making it difficult for the character-level model to collect enough information. With larger datasets, this model will be able to perform significantly better\cite{zhang}. This can be illustrated from Figure \ref{Fig:comparison} that with a larger number of samples, the Char-CNN model raises steeply and performs competitively with the other models. In terms of the number of parameters, character level model is much superior to its word-level counterparts. The embedding vectors for the word level models is of size $embedding\ vector\;*\;vocabulary\ size$. i.e. 300 * 60000. On the other hand, the character embedding matrix is of size 253*253 given that we initially used one-hot vectors. This size can also be reduced to as low as 253*16 as were done in some research\cite{googlebrain}. Another thing to consider is the time it takes to train the models. For the word embedding models, a pure CNN does not work satisfactorily, so an LSTM layer had to be added to add sequential information in the model. This improves accuracy with the cost of taking more time to train, around 15-20 minutes. On the other hand, the character-level model works significantly well with only using convolutional layers taking less than 2 minutes to train. This effect of training time become largely magnified on large-scale cases, making the word-level model unfit for light-weight devices. As stated in the paper\cite{zhang}, ConvNets with character embedding can completely replace words and work even without any semantic meanings. Which means that convolutional layers can extract whatever information necessary for author attribution, given enough data.\newline
To illustrate the need of pre-trained character embeddings, we see from \ref{comptable} that using a pre-trained embedding increases the accuracy across datasets and the different number of authors, regardless of the amount of data for each author. Which shows that these naively learned embeddings contain valuable information that can be easily applied to various tasks of the language, including author attribution, and increase the performance a few degrees. These numerical representations of character contain information about morphology and the syntax of the language among other things. Therefore such embedding can be learned from any task and applied to other tasks as a form of transfer learning, given the alphabet remains the same.   

\section{Conclusion}
So far no work has been done to evaluate the usefulness of character embeddings for classification task in Bangla language. We attempt to fill this gap and compare character embeddings with word embeddings showing that character embeddings perform almost as good as the best word embedding model. But besides accuracy, character level classification has a greater hand in terms of memory, time and number of parameters. Considering the small size of our datasets, we hope to have improved performance with larger datasets, as is the case for character level ConvNets\cite{zhang}.  Besides such network also work better with non-curated texts, which are hard for word-level embeddings to capture, thus more applicable to real-life scenarios. Furthermore, we analyzed the importance of pre-trained character embedding for author attribution and showed that pre-training can result in better performances. Since very large corpus is not available in Bangla language yet, we must come up with solutions that tackle attribution tasks sufficiently well even with little data. Therefore our future works include the combination of both character and word level embeddings to perform attribution task, in an attempt to combine the power of both types of embeddings. More advanced levels of transfer learning can also be performed by using language models in place of embeddings before classification. Language models and embeddings can also be combined to give greater generalization for Bangla language.

\vspace{12pt}

\bibliography{main} 
\bibliographystyle{ieeetr}
\end{document}